\documentclass[a4paper,conference]{IEEEtran}
\usepackage{graphicx}
\usepackage{subcaption}
\usepackage{amssymb}
\usepackage{amsmath}
\usepackage{color}
\usepackage{latexsym}
\usepackage{float}
\usepackage[compress,noadjust]{cite}
\usepackage{multirow}
\usepackage[bookmarks=false]{hyperref}
\usepackage{hyperref}
\usepackage{algorithm, algorithmic}
\usepackage{tabularx}
\newcommand{\minisection}[1]{\vspace{0.04in} \noindent {\bf #1}\ \ }

\renewcommand{\thefootnote}{\fnsymbol{footnote}}
\begin{document}
%
\title{Rotate your Networks: Better Weight Consolidation and Less Catastrophic Forgetting}

\author{\IEEEauthorblockN{Xialei Liu$^*$, Marc Masana$^{*}$, Luis Herranz,
Joost Van de Weijer, Antonio M. L\'opez}
\IEEEauthorblockA{ Computer Vision Center, UAB \\Barcelona, Spain\\
Email: \{xialei, mmasana, lherranz, joost, antonio\}@cvc.uab.es}
\and
\IEEEauthorblockN{Andrew D. Bagdanov}
\IEEEauthorblockA{MICC, University of Florence \\Florence, Italy\\
Email: andrew.bagdanov@unifi.it}}

\maketitle

\begin{abstract}
  In this paper we propose an approach to avoiding catastrophic
  forgetting in sequential task learning scenarios. Our technique is
  based on a network reparameterization that approximately
  diagonalizes the Fisher Information Matrix of the network
  parameters. This reparameterization takes the form of a factorized
  rotation of parameter space which, when used in conjunction with
  Elastic Weight Consolidation (which assumes a diagonal Fisher Information
  Matrix), leads to significantly better performance on lifelong
  learning of sequential tasks. Experimental results on the MNIST,
  CIFAR-100, CUB-200 and Stanford-40 datasets demonstrate that we
  significantly improve the results of standard elastic weight
  consolidation, and that we obtain competitive results when compared
  to the state-of-the-art in lifelong learning without
  forgetting.
  \footnotetext[1]{Both authors contributed equally.}
\end{abstract}


\section{Introduction}

Neural networks are very effective models for a variety of computer
vision tasks. In general, during training these networks are presented
with examples from all tasks they are expected to perform. In a
lifelong learning setting, however, learning is considered as a
sequence of tasks to be learned~\cite{silver2002task}, which is more
similar to how biological systems learn in the real world. In this
case networks are presented with \emph{groups} of tasks, and at
any given moment the network has access to training data from only
\emph{one} group. The main problem which systems face in such settings
is \emph{catastrophic forgetting}: while adapting
network weights to new tasks, the network forgets the previously learned
ones~\cite{mccloskey1989catastrophic}.

There are roughly two main approaches to lifelong learning (which has
seen increased interest in recent years). The first group of methods
stores a small subset of training data from previously learned
tasks. These stored exemplars are then used during training of new
tasks to avoid forgetting the previous
ones~\cite{rebuffi2016icarl,lopez2017gradient}. The second type of
algorithm instead avoids storing any training data from previously
learned tasks. A number of algorithms in this class are based on
Elastic Weight Consolidation
(EWC)~\cite{kirkpatrick2017overcoming,lee2017overcoming,zenke2017continual},
which includes a regularization term that forces parameters of the
network to remain close to the parameters of the network trained for
the previous tasks. In a similar vein, Learning Without Forgetting
(LWF)~\cite{li2016learning} regularizes \emph{predictions} rather than
weights. Aljundi
et al.~\cite{aljundi2016expert} learns a representation for each task and use a set of gating autoencoders to decide which expert to use at testing time.

EWC is an elegant approach to selective regularization of network
parameters when switching tasks. It uses the Fisher Information Matrix
(FIM) to identify directions in feature space critical to performing
already learned tasks (and as a consequence also those directions in
which the parameters may move freely without forgetting learned
tasks). However, EWC has the drawback that it assumes the Fisher
Information Matrix to be diagonal -- a condition that is almost never
true.

\begin{figure}
\centering
\includegraphics[width=0.99\columnwidth]{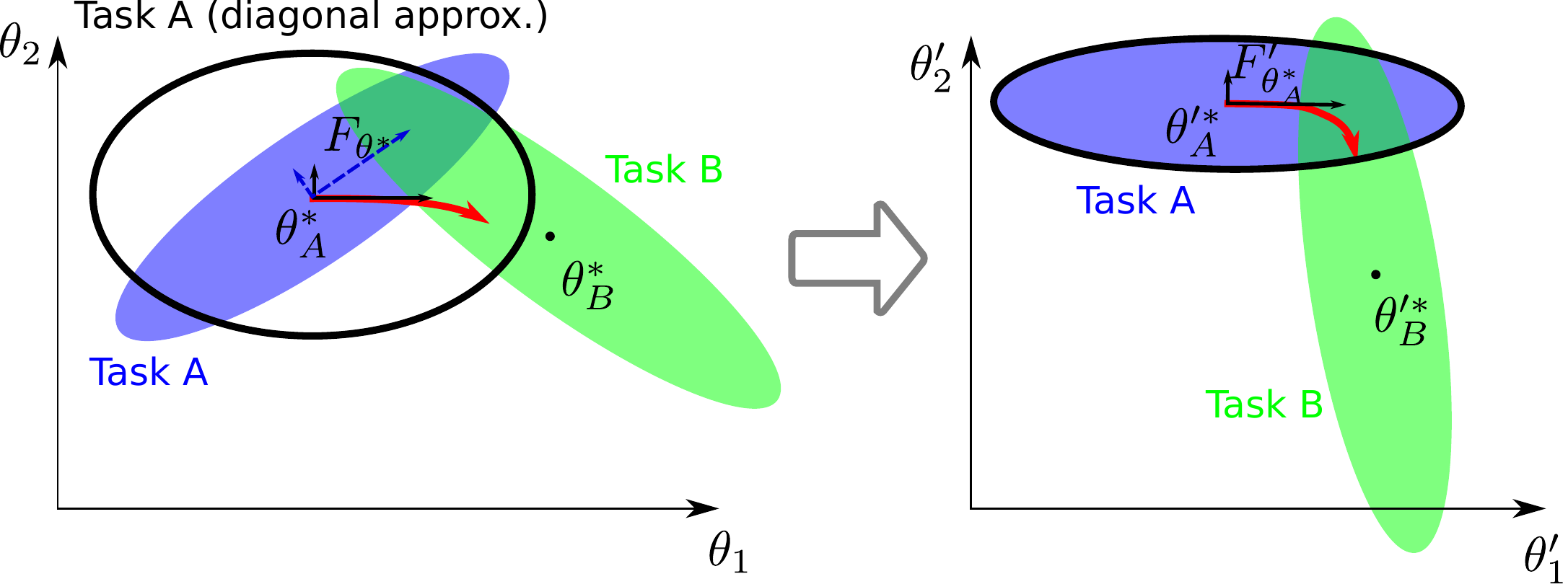}
\caption{Sequential learning in parameter space, illustrating the
  optimal model parameters for tasks A and B and regions with low
  forgetting. The red line indicates the learning path for task B from
  the previously learned solution for task A.  \textbf{Left}: the
  diagonal approximation (black ellipse) of the Fisher Information
  Matrix can be poor and steer EWC in directions that will forget task
  A. \textbf{Right}: after a suitable reparameterization (i.e., a
  rotation) the diagonal approximation is better and EWC can avoid
  forgetting task A.}
\label{fig:motivation_rotation}
\end{figure}

In this paper we specifically address this diagonal assumption made by
the EWC algorithm. If the FIM is not diagonal, EWC can fail to prevent
the network from straying away from ``good parameter space'' (see
Fig.~\ref{fig:motivation_rotation}, left). Our method is based on
\emph{rotating} the parameter space of the neural network in such a
way that the output of the forward pass is unchanged, but the FIM
computed from gradients during the backward pass is approximately
diagonal (see Fig.~\ref{fig:motivation_rotation}, right). The result
is that EWC in this rotated parameter space is significantly more
effective at preventing catastrophic forgetting in sequential task
learning problems. An extensive experimental evaluation on a variety
of sequential learning tasks shows that our approach significantly
outperforms standard elastic weight consolidation. 

The rest of the paper is organized as follows. In the next section we
review the EWC learning algorithm and in Section~\ref{sec:method} we
describe the approach to rotating parameter space in order to better
satisfy the diagonal requirement of EWC. We give an extensive
experimental evaluation of the proposed approach in
Section~\ref{sec:experiments} and conclude with a discussion of our
contribution in Section~\ref{sec:conclusions}.

\section{Related work}
\label{sec:related}
\renewcommand{\thefootnote}{\arabic{footnote}}
An intuitive approach to avoiding forgetting for sequential task
learning is to retain a portion of training data for each task. The
iCaRL method of Rebuffi et al.~\cite{rebuffi2016icarl} is based on
retaining exemplars which are rehearsed during the training of new
tasks. Their approach also includes a method for exemplar herding
which ensures that the class mean remains close even when the number
of exemplars per class is dynamically varied. A recent paper proposed
the Gradient Episodic Memory model~\cite{lopez2017gradient}. Its
main feature is an episodic memory storing a subset of the observed
examples. However, these examples are not directly rehearsed but
instead are used to define inequality constraints on the loss that
ensure that it does not increase with respect to previous tasks. A
cooperative dual model architecture consisting of a deep generative
model (from which training data can be sampled) and a task solving
model is proposed in~\cite{shin2017continual}.

Learning without Forgetting in~\cite{li2016learning} works without
retaining training data from previous tasks by regularizing the
network output on new task data to not stray far from its original
output. Similarly, the lifelong learning approach described
in~\cite{rannen2017encoder} identifies informative features using
per-task autoencoders and then regularizes the network to preserve
these features in its internal, shared-task feature representation
when training on a new task. Elastic Weight Consolidation
(EWC)~\cite{kirkpatrick2017overcoming,lee2017overcoming,zenke2017continual}
includes a regularization term that forces important parameters of the
network to remain close to the parameters of the network trained for
the previous tasks. The authors of~\cite{serra2018overcoming} propose
a task-based hard attention mechanism that preserves information about
previous tasks without affecting the current task's learning. Aljundi
et al.~\cite{aljundi2016expert} learn a representation for each task and use a set of gating autoencoders to decide which expert to use at testing time. 

Another class of learning methods related to our approach are those
based on the \emph{Natural Gradient Descent
  (NGD)}~\cite{amari1998natural,pascanu2013revisiting}. The NGD uses
the Fisher Information Matrix (FIM) as the natural metric in parameter
space. Candidate directions are projected using the inverse FIM and
the best step (in terms of decreasing loss) is taken. The Projected
NGD algorithm estimates a network reparamaterization online that
whitens the FIM so that vanilla Stochastic Gradient Descent (SGD) is
equivalent to NGD~\cite{desjardins2015natural}. The authors
of~\cite{lafond2017diagonal} show how the natural gradient can be used
to explain and classify a range of adaptive stepsize strategies for
training networks. In~\cite{grosse2016kronecker} the authors propose
an approximation to the FIM for convolutional networks and show that
the resulting NGD training procedure is many times more efficient than
SGD.

\section{Elastic weight consolidation}
\label{sec:ewc}

Elastic Weight Consolidation (EWC) addresses the problem of
catastrophic forgetting in continual and sequential task learning in
neural
networks~\cite{kirkpatrick2017overcoming,mccloskey1989catastrophic}. In
this section we give a brief overview of EWC and discuss some of its
limitations.

\subsection{Overview}

The problem addressed by EWC is that of learning the $K$-th task in a
network that already has learned $K\!-\!1$ tasks. The main challenge is to
learn the new task in a way that prevents \textit{catastrophic
  forgetting}. Catastrophic forgetting occurs when new learning
interferes catastrophically with prior learning during sequential
neural network training, which causes the network to partially or
completely forget previous tasks~\cite{mccloskey1989catastrophic}.

The final objective is to learn the optimal set of parameters
$\theta_{1:K}^*$ of a network given the previous ones
$\theta_{1:K-1}^*$ learned from the previous $K\!-\!1$ tasks.\footnote{We
  adapt the notation from \cite{lee2017overcoming}
  and~\cite{desjardins2015natural}.} Each of the $K$ tasks consists of
a training dataset
$\mathcal{D}_k = \left(\mathcal{X}_k,\mathcal{Y}_k\right)$ with
samples $x\in \mathcal{X}_k$ and labels $y\in \mathcal{Y}_k$ (and for
previous tasks
$\mathcal{D}_{1:K-1} =
\left(\mathcal{X}_1,\ldots,\mathcal{X}_{K-1},\mathcal{Y}_1,\ldots,\mathcal{Y}_{K-1}\right))$. We
are interested in the configuration $\theta$ which maximizes the
posterior $
	p_{1:K} \equiv p\left(\theta \vert \mathcal{D}_{1:K} \right)
$. 

EWC \cite{kirkpatrick2017overcoming} uses sequential Bayesian estimation \cite{lee2017overcoming} to factorize $p_{1:K}$ as:
\begin{eqnarray}
\log p_{1:K} \!\!\! &=& \!\!\! \log p\left(\mathcal{Y}_K\vert \mathcal{X}_K; \theta\right) + \log p\left(\theta \vert \mathcal{D}_{1:K-1} \right)\\
& & \!\!\! -\log p\left(\mathcal{D}_{K} \vert \mathcal{D}_{1:K-1} \right) + C\nonumber,
\label{eq:model_1_K_2}
\end{eqnarray} 
where $p_{1:K-1}$ is the prior model from previous tasks,
$p_K \equiv p\left( \theta \vert \mathcal{X}_K, \mathcal{Y}_K \right)$
is the posterior probability corresponding to the new task, and $C$ is
a constant factor that can be ignored for training.

Since calculating the true posterior is intractable, EWC uses the Laplace
approximation to approximate the posterior with a
Gaussian:

\begin{eqnarray} 
  \log p_{1:K} \!\!\!\! &\approx& \!\!\!\! \log p\left(\mathcal{Y}_K\vert \mathcal{X}_K; \theta\right) \label{eq:EWC_full} \\
              &&  - \frac{\lambda}{2} \left( \theta - \theta_{1:K-1}^* \right)^\intercal \tilde{F}_{1:K-1} \left( \theta - \theta_{1:K-1}^* \right) + C' \nonumber \\
              \!\!\!\! &\approx& \!\!\!\! \log p\left(\mathcal{Y}_K\vert \mathcal{X}_K; \theta\right) \label{eq:EWC_diag} \\
 && - \frac{\lambda}{2} \sum_i \tilde{F}_{1:K-1,ii}\! \left(\theta_i \! - \theta_{1:K-1,i}^* \right)^2 + C',
\nonumber \label{eq:EWC_sequential} 
\end{eqnarray} 
where $\tilde{F}_{1:K-1}$ is the Fisher Information Matrix (FIM) that
approximates the inverse of the covariance matrix at
$\theta_{1:K-1}^*$ used in the Laplace approximation of $\log p_{1:l}$
in (\ref{eq:EWC_full}), and the second approximation in
(\ref{eq:EWC_diag}) assumes $\tilde{F}$ is diagonal -- a common
assumption in practice -- and thus that the quadratic form in
(\ref{eq:EWC_full}) can be replaced with scaling by the diagonal
entries $\tilde{F}_{1:K-1,ii}$ of the FIM at $\theta_{1:K-1}^*$.

The FIM of the true distribution
$p\left(y\vert x; \theta\right)$ is estimated as:
\begin{eqnarray}
  F_\theta \!\!\!\! &=& \!\!\!\! \mathbb{E}_{x\sim \pi} \left\{ \mathbb{E}_{y\sim p\left(y\vert x; \theta\right)}\left[\left (\frac{\partial \log p}{\partial \theta}\right) \!\! \left(\frac{\partial \log p}{\partial \theta}\right )^{\intercal}\right] \right\} \label{eq:FIM_firstorder}\\
          \!\!\!\! &=& \!\!\!\! \mathbb{E}_{x\sim \pi} \left\{ \mathbb{E}_{y\sim p\left(y\vert x; \theta\right)}\left[\frac{\partial^2 \log p}{\partial \theta^2} \right] \right\},
\label{eq:FIM_secondorder}
\end{eqnarray}
where $\pi$ is the empirical distribution of a training set
$\mathcal{X}$.

This definition of the FIM as the second derivatives of the
log-probability is key to understanding its role in preventing
forgetting. Once a network is trained to a configuration $\theta^*$,
the FIM $F_{\theta^*}$ indicates how prone each dimension in the
parameter space is to causing forgetting when gradient descent updates
the model to learn a new task: it is preferable to move along
directions with low Fisher information, since $\log p$ decreases
slowly (the red line in Fig.~\ref{fig:motivation_rotation}). EWC uses
$F_{\theta^*}$ in the regularization term of (\ref{eq:EWC_full}) and
(\ref{eq:EWC_diag}) to penalize moving in directions with higher
Fisher information and which are thus likely to result in forgetting
of already-learned tasks.  EWC uses different regularization terms per
task~\cite{kirkpatrick2017overcoming}. Instead of using
(\ref{eq:EWC_diag}), we only use the FIM from the previous task
because we found it works better, requires storage of only one FIM,
and better fits the sequential Bayesian perspective (as shown
in~\cite{huszar2018note}).

\subsection{Limitations of EWC}
Assuming the FIM to be diagonal is a common practice in the Laplace
approximation for two reasons. First, the number of parameters is
reduced from $O(N^2)$ to $O(N)$, where $N$ is the number of elements
in parameter space $\theta$, so a diagonal matrix is much more
efficient to compute and store. Additionally, in many cases the
required matrix is the inverse of the FIM (for example in NGD methods~\cite{amari1998natural}), which is significantly
simpler and faster to compute for diagonal matrices.

However, in the case of sequential learning of new tasks with EWC, the
diagonal FIM assumption might be unrealistic -- at least in the
original parameter space. On the left of
Fig.~\ref{fig:motivation_rotation} is illustrated a situation where a
simple Gaussian distribution (solid blue ellipse) is approximated
using a diagonal covariance matrix (black ellipse). By rotating the
parameter space so that $\frac{\partial \log p}{\partial \theta}$ are
aligned (on average) with the coordinate axes
(Fig.~\ref{fig:motivation_rotation}, right), the diagonal assumption
is more reasonable (or even true in the case of a Gaussian
distribution). In this rotated parameter space, EWC is better able to
optimize the new task while not forgetting the old one.

\begin{figure}[tb]
	\begin{centering}
  \begin{subfigure}[b]{0.47\columnwidth}
    \includegraphics[width=\textwidth]{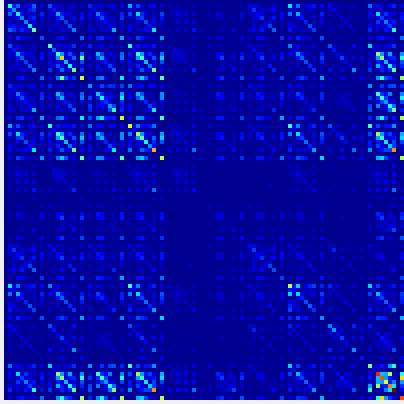}
    \caption{Full}
    \label{fig:FisherMatrix_ori}
  \end{subfigure}
  %
    \begin{subfigure}[b]{0.47\columnwidth}
    \includegraphics[width=\textwidth]{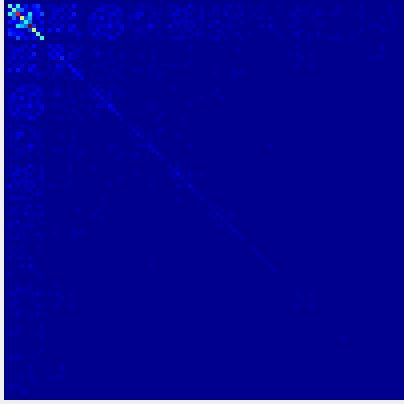}
    \caption{Rotated}
    \label{fig:FisherMatrix_rot}
  \end{subfigure}

    \end{centering}
    \caption{\label{fig:FisherMatrix} Fisher Information Matrix: (a) original and (b) rotated using the proposed technique. The range is color coded and normalized for better visualization.}
\end{figure}

The example in Fig.~\ref{fig:FisherMatrix_ori} shows the FIM obtained
for the weights in the second layer of a multilayer perceptron trained
on MNIST~\cite{srivastava2013compete} (specifically, four dense layers
with 784, 10, 10, and 10
neurons). The matrix is clearly non-diagonal, so the diagonal
approximation misses significant correlations between weights and this
may lead to forgetting when using the diagonal approximation. The
diagonal only retains 40.8\% of the energy of the full matrix in this
example.

\section{Rotated elastic weight consolidation}
\label{sec:method}

Motivated by the previous observation, we aim to find a
reparameterization of the parameter space $\theta$. Specifically, we
desire a reparameterization which does not change the feed-forward
response of the network, but that better satisfies the assumption of
diagonal FIM. After reparameterization, we can assume a diagonal FIM
which is efficiently estimated in that new parameter space. Finally,
minimization by gradient descent on the new task is also performed in
this new space.

One possible way to obtain this reparameterization is by computing a
rotation matrix using the Singular Value Decomposition (SVD) of
(\ref{eq:FIM_firstorder}). Note that this decomposition is performed in the parameter space.  Unfortunately, this approach
has three problems. First, the SVD is extremely expensive to compute on very large matrices. Second, this rotation ignores the sequential
structure of the neural network and would likely catastrophically
change the feed-forward behavior of the network. Finally, we do not
have the FIM in the first place.

\subsection{Indirect rotation}
In this section we show how rotation of fully connected and
convolutional layer parameters can be applied to obtain a network for
which the assumption of diagonal FIM is more valid. These rotations
are chosen so as to not alter the feed-forward response of the
network.

For simplicity, we first consider the case of a single fully-connected
layer given by the linear model $\mathbf{y}=W\mathbf{x}$, with
input $\mathbf{x} \in \mathbb{R}^{d_1}$, output
$\mathbf{y} \in \mathbb{R}^{d_2}$ and weight matrix
$W\in \mathbb{R}^{d_2 \times d_1}$. In this case $\theta=W$, and to
simplify the notation we use
$L=\log p\left(\mathbf{y}\vert\mathbf{x};W\right)$. Using (\ref{eq:FIM_firstorder}),
the FIM in this simple linear case is (after applying the chain rule):
\begin{eqnarray}
F_W &=& \mathbb{E}_{\substack{x\sim \pi \\ y\sim p }}\left[\left (\frac{\partial L}{\partial
  \mathbf{y}}\frac{\partial \mathbf{y}}{\partial W}\right ) \left
  (\frac{\partial L}{\partial \mathbf{y}} \frac{\partial
  \mathbf{y}}{\partial W}\right )^{\intercal} \right] \nonumber \\
 &=& \mathbb{E}_{p\sim \pi}\left[\left (\frac{\partial L}{\partial \mathbf{y}}\right) \mathbf{x}\mathbf{x}^{\intercal} \left (\frac{\partial L}{\partial \mathbf{y}}\right )^{\intercal} \right]. \label{eq:FIM_W_joint}
\end{eqnarray}

If we assume that $\frac{\partial L}{\partial \mathbf{y}}$ and $\mathbf{x}$ are independent random variables we can factorize (\ref{eq:FIM_W_joint}) as done in~\cite{desjardins2015natural}:
\begin{equation}
F_W=\mathbb{E}_{\substack{x\sim \pi \\ y\sim p }}\left[\left (\frac{\partial L}{\partial \mathbf{y}}\right) \left (\frac{\partial L}{\partial \mathbf{y}}\right )^{\intercal}\right] \mathbb{E}_{x\sim \pi}\left[ \mathbf{x}\mathbf{x}^{\intercal}  \right], \label{eq:FIM_W_split3}
\end{equation}
which indicates that we can approximate the FIM using two independent
factors that only depend on the backpropagated gradient at the output
$\frac{\partial L}{\partial \mathbf{y}}$ and on the input
$\mathbf{x}$, respectively. This result also suggests that there may
exist a pair of rotations of the input and the output, respectively,
that lead to a rotation of $F_W$ in the parameter space $W$.

\begin{figure}
	\centering
	\includegraphics[width=0.99\columnwidth]{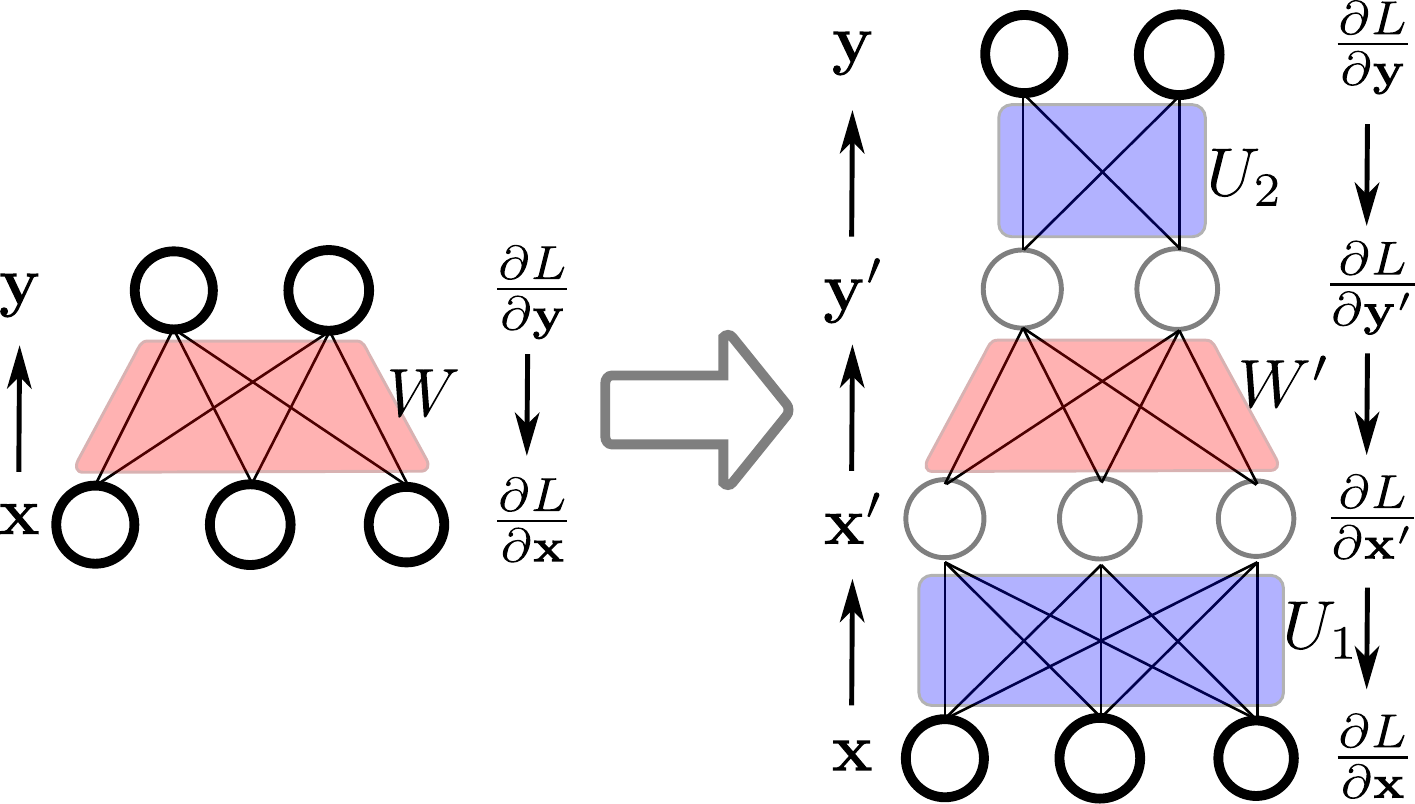}
	\caption{Reparameterization of a linear layer $W$ as $W'$ using two additional linear layers $U_1$ and $U_2$.}
	\label{fig:layer_rotation}
\end{figure}

In fact, these rotation matrices can be obtained as
$U_1\in\mathbb{R}^{d_1 \times d_1}$ and $U_2\in\mathbb{R}^{d_2 \times d_2}$
from the following SVD decompositions:
\begin{eqnarray}
\mathbb{E}_{x\sim \pi}\left[ \mathbf{x}\mathbf{x}^{\intercal}  \right] &=& U_1S_1V_1^\intercal\label{eq:FIM_W_split} \\
\mathbb{E}_{\substack{x\sim \pi \\ y\sim p }}\left[\left (\frac{\partial L}{\partial \mathbf{y}}\right) \left (\frac{\partial L}{\partial \mathbf{y}}\right )^{\intercal}\right] &=& U_2S_2V_2^\intercal \label{eq:FIM_W_split2}
\end{eqnarray}
Since both rotations are local, i.e. they are applied to a single layer, they can be integrated in
the network architecture as two additional (fixed) linear layers
$\mathbf{x}'=U_1\mathbf{x}$ and $\mathbf{y}=U_2\mathbf{y}'$ (see
Fig.~\ref{fig:layer_rotation}). 

The new, rotated weight matrix is then:

\begin{equation}
W'=U_2^\intercal W U_1^\intercal.
\end{equation}
Thus, that the forward passes of both networks in
Fig.~\ref{fig:layer_rotation} is equivalent since $U_2 W' U_1 = W$. In
this way, the sequential structure of the network is not broken, and
the learning problem is equivalent to learning the parameters of $W'$
for the new problem $\mathbf{y}'=W'\mathbf{x}'$. The use of layer
decomposition with SVD was also investigated in the context of network
compression~\cite{li2013,marc2017domain}. However this SVD analysis
was based on layer weight matrices and the original network layer was
only decomposed into two new layers.

The training procedure is exactly the same as in
(\ref{eq:model_1_K_2}), but using $\mathbf{x}'$, $\mathbf{y}'$ and
$W'$ instead of $\mathbf{x}$, $\mathbf{y}$ and $W$ to estimate the FIM
and to learn the weights. The main difference is that the approximate
diagonalization of $W$ in $W'$ will be more effective in preventing
forgetting using EWC. Fig.~\ref{fig:FisherMatrix_rot} shows the
resulting matrix after applying the proposed rotations in the previous
example. Note that most of the energy concentrates in fewer
weights and it is also better conditioned for a diagonal approximation
(in this case the diagonal retains 74.4\% of the energy).

Assuming a block-diagonal FIM, the extension to multiple layers is
straightforward by applying the same procedure layer-wise using the
corresponding inputs instead of $\mathbf{x}$ and backpropagated
gradients to the output of the layer as
$\frac{\partial L}{\partial \mathbf{y}}$ (that is, estimating the FIM
for each layer, and computing layer-specific $U_1$, $W'$ and
$U_2$). In Algorithm~\ref{table:algorithm} we describe the
reparameterization used in the training process.

\subsection{Extension to convolutional layers}

\begin{figure}
  \centering
  \includegraphics[width=0.8\columnwidth]{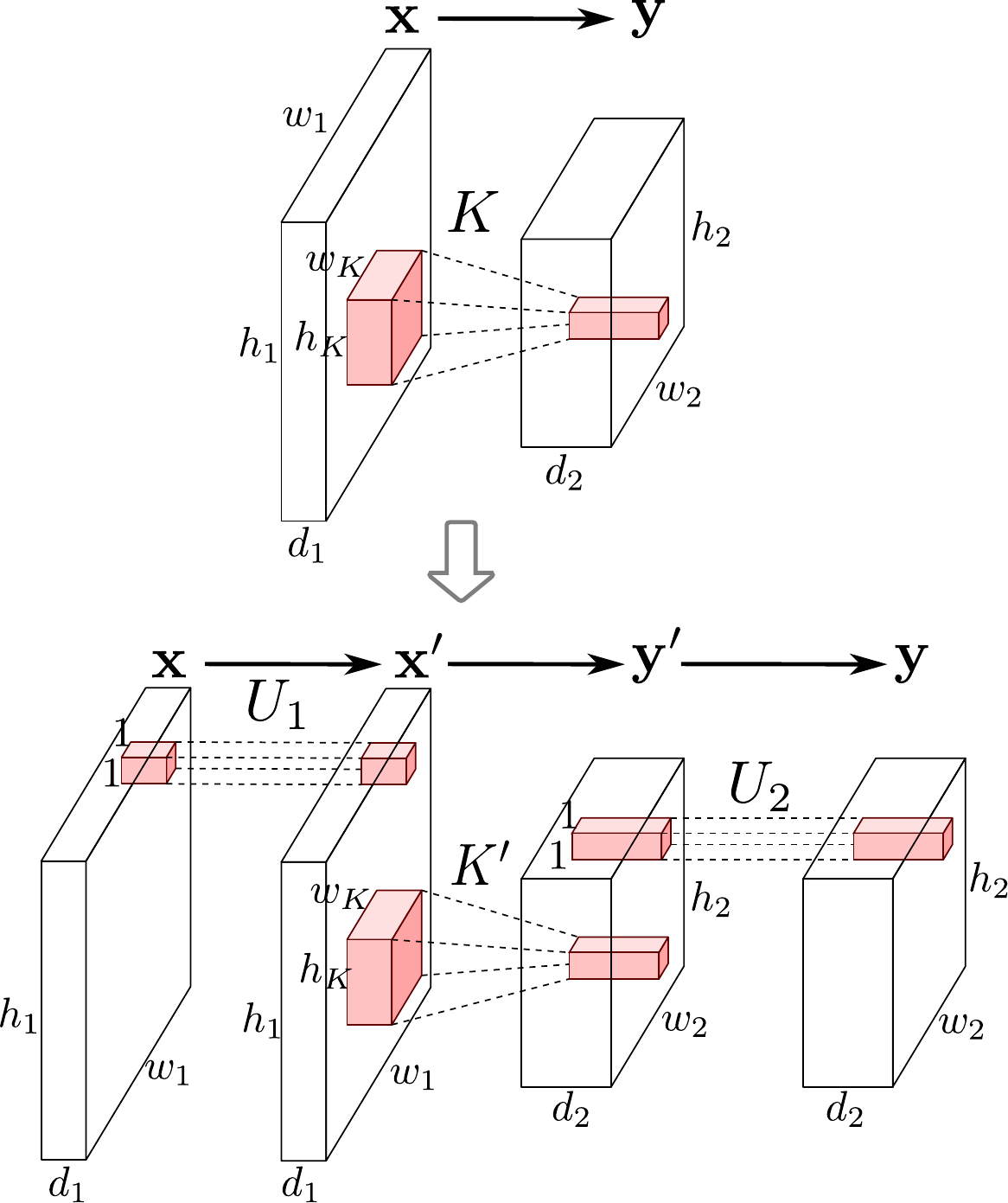}
  \caption{Reparameterization of a convolutional layer $K$ as $K'$
    using two additional $1\times 1$ convolutional layers $U_1$ and
    $U_2$ as rotations.}
  \label{fig:layer_rotation_conv}
\end{figure}

The method proposed in the previous section for fully connected layers
can be applied to convolutional layers with only slight
modifications. The rotations are performed by adding two
additional $1\times 1$ convolutional layers (see
Fig.~\ref{fig:layer_rotation_conv}).

Assume we have an input tensor
$\mathbf{x}\!\in\!\mathbb{R}^{w_1 \times h_1 \times d_1}$, a kernel
tensor $K\!\in\!\mathbb{R}^{w_k \times h_k \times d_1 \times d_2}$, and
that the corresponding output tensor is
$\mathbf{y}\!\in\!\mathbb{R}^{w_2 \times h_2 \times d_2}$. For
convenience let the mode-3 fiber\footnote{\label{tensor_notations}A mode-$i$ fiber of a tensor is defined as the vector obtained by fixing all its indices but $i$. A slice of a tensor is a matrix obtained by fixing all its indices but two. See \cite{kolda2009tensor} for more details.}
of $\mathbf{x}$ be
$\mathbf{x}_{l,m}\!=\!\left( x_{l,m,i} \vert i=1,\ldots,d_1
\right)^\intercal$ and of the output gradient tensor
$\frac{\partial L}{\partial \mathbf{y}}$ as
$\mathbf{z}_{l,m} = \left( \left (\frac{\partial L}{\partial
      \mathbf{y}}\right)_{l,m,i} \vert i=1,\ldots,d_2
\right)^\intercal$. Note that each $\mathbf{x}_{l,m}$ and
$\mathbf{z}_{l,m}$ are $d_1$-dimensional and $d_2$-dimensional
vectors, respectively. Now we can compute the self-correlation
matrices averaged over all spatial coordinates as:
\begin{eqnarray}
X = \frac{1}{w_1h_1} \sum _{ l=1 }^{ w_1 } \sum _{ m=1 }^{ h_1 } \mathbf{x}_{l,m} \mathbf{x}_{l,m}^{\intercal}
\\
Z = \frac{1}{w_2h_2} \sum _{ l=1 }^{ w_2 } \sum _{ m=1 }^{ h_2 } \mathbf{z}_{l,m} \mathbf{z}_{l,m}^{\intercal},
\end{eqnarray}
and compute the decompositions of (\ref{eq:FIM_W_split}) and (\ref{eq:FIM_W_split2}) as
\begin{eqnarray}
\mathbb{E}_{x\sim \pi}\left[ X \right] &=& U_1S_1V_1^\intercal
\\
\mathbb{E}_{\substack{x\sim \pi \\ y\sim p }}\left[ Z \right] &=& U_2S_2V_2^\intercal.
\end{eqnarray}

We define $K_{l,m}=\left( k_{l,m,i,j} \vert i=1,\ldots,d_1, j=1,\ldots,d_2 \right)$, which is a slice\textsuperscript{\ref{tensor_notations}} of the kernel tensor $K$. The rotated slices are then obtained as:
\begin{eqnarray}
K'_{l,m} = U_2^\intercal K_{l,m} U_1^\intercal.
\end{eqnarray}
and the final rotated kernel tensor
$K'\in \mathbb{R}^{w_k \times h_k \times d_1 \times d_2}$ is obtained
simply by tiling all slices $K'_{l,m}$ computed for
every $l$ and $m$.

\begin{algorithm}[tb]
\caption{: Incremental task learning. }
\label{table:algorithm}
\begin{algorithmic}
\STATE\textbf{Input:} $\mathcal{D}_k =
\left(\mathcal{X}_k,\mathcal{Y}_k\right)$, training samples in per-class sets. \\
\STATE\textbf{Require:} Initial parameters $\Theta_0$, total number of
tasks $K$ \\
\textbf{for} $k=1,\ldots,K$ \\
\ \ \ \ \textbf{if} $k$=1 \\
\ \ \ \ \ \ \ \ $\Theta$  $\leftarrow$ UPDATE($\left(\mathcal{X}_k,
  \mathcal{Y}_k\right)$, $\Theta_0$, $0$) \\
\ \ \ \ \textbf{else}
\vspace{-0.1in}
\begin{alignat*}{3}
& F_\Theta  && \leftarrow && \mbox{ COMPUTE\_FIM}(\left(\mathcal{X}_{k-1}, \mathcal{Y}_{k-1}\right), \Theta^R) \\
& \Theta^R && \leftarrow && \mbox{ UPDATE}(\left(\mathcal{X}_k, \mathcal{Y}_k\right), \Theta^R, F_\Theta) \\
& \Theta   && \leftarrow && \mbox{ COMBINE}(\Theta^R) \\
\end{alignat*} \\
\vspace{-0.25in}
\ \ \ \ $\Theta^R$  $\leftarrow$ ROTATE($\Theta$) \\
\textbf{end for}

\end{algorithmic}
\vspace{0.05in}
\hrule
\vspace{0.05in}
$\mbox{UPDATE}(\left(\mathcal{X}_k, \mathcal{Y}_k\right), \Theta^R,
F)$ above fits the model for task $k$ using EWC, with rotated
parameters $\Theta^R$, and FIM $F$ ($0$ is the zero matrix). $\mbox{COMBINE}(\Theta^R)$ fuses
current parameters before computation of new rotated parameters for the coming tasks.  
\end{algorithm}

\section{Experimental results}
\label{sec:experiments}

In this section we report on a number of experiments comparing our
approach to EWC~\cite{kirkpatrick2017overcoming} and other baselines.\footnote{Code available at:
  \url{https://github.com/xialeiliu/RotateNetworks}}

\subsection{Experimental settings}
\begin{table*}[tb]
\begin{center}
\caption{Ablation study on Disjoint MNIST when $T\!=\!2$. Numbers in
  \textbf{bold} indicate the best performing configuration of each
  method. All versions of R-EWC that rotate fully-connected layers
  significantly outperform EWC.}
\label{tab:exp1_mnist_lenet}
\resizebox{0.95\textwidth}{!}{%
\begin{tabular}{rcccccccccc}
\hline
 & \multicolumn{2}{c}{$\lambda=1$} & \multicolumn{2}{c}{$\lambda=10$} & \multicolumn{2}{c}{$\lambda=100$} & \multicolumn{2}{c}{$\lambda=1000$} & \multicolumn{2}{c}{$\lambda=10000$} \\
 & Task 1 & Task 2 & Task 1 & Task 2 & Task 1 & Task 2 & Task 1 & Task 2 & Task 1 & Task 2 \\ \hline\hline
FT                  & 6.1 & 97.6 & 6.1 & 97.6 & 6.1 & 97.6 & 6.1 & 97.6 & 6.1 & 97.6 \\
EWC~\cite{kirkpatrick2017overcoming} & 66.8 & 90.9 & 75.3 & 95.6 & \textbf{85.8} & \textbf{92.8} & 78.4 & 93.7 & 81.0 & 88.8 \\ \hline
R-EWC - conv only    & 62.7 & 89.2 & 67.5 & 96.1 & 80.4 & 91.4 & \textbf{84.7} & \textbf{93.1} & 75.5 & 93.7 \\
R-EWC - fc only      & 78.9 & 95.3 & 79.0 & 95.8 & 87.4 & 93.5 & 93.0 & 82.3 & \textbf{94.3} & \textbf{88.0} \\ 
R-EWC - all          & 77.2 & 96.7 & \textbf{91.7} & \textbf{91.2} & 86.9 & 95.9 & 96.3 & 81.1 & 92.1 & 86.0 \\
R-EWC - all no last  & 71.5 & 91.8 & 84.9 & 97.0 & \textbf{91.6} & \textbf{94.5} & 94.6 & 88.4 & 97.9 & 79.4 \\ \hline
\end{tabular}%
}

\end{center}
\end{table*}

\minisection{Datasets.} We evaluate our method on two small datasets
and three fine-grained classification datasets: MNIST~\cite{srivastava2013compete}, CIFAR-100~\cite{krizhevsky2009learning}, CUB-200 Birds~\cite{WahCUB_200_2011} and
Stanford-40 Actions~\cite{yao2011human}. Each dataset is equally divided into $T$
groups of classes, which are seen as the sequential tasks to learn. In
the case of the CUB-200 dataset, we crop the bounding boxes from the
original images and resize them to 224$\times$224. For
Actions, we resize the input images to 256$\times$256, then take
random crops of size 224$\times$224. We perform no data augmentation
for the MNIST and CIFAR-100 datasets.

\minisection{Training details.}  We chose the LeNet~\cite{lecun1998gradient} and
VGG-16~\cite{Simonyan2014vgg} network architectures, which are
slightly modified due to the requirements of different datasets. For
MNIST, we add 2$\times$2 padding to the original 28$\times$28 images
to obtain 32$\times$32 input images for LeNet. LeNet is trained from scratch, 
while for CIFAR-100 the
images are passed through the VGG-16~\cite{Simonyan2014vgg} network
pre-trained on ImageNet, which has been shown to perform well when
changing to other domains. The input images are 32$\times$32$\times$3
and this provides a feature vector of 1$\times$1$\times$512 at the end
of the \textit{pool5} layer. We use those feature vectors as an input
to a classification network consisting of 3 fully-connected layers of
output dimensions of 256, 256 and 100, respectively. For the two
fine-grained datasets, we fine-tune from the pre-trained model on
ImageNet. To save memory and limit computational complexity, we add a
global pooling layer after the final convolutional layer of
VGG-16. The fully-connected layers used for our experiments are of
size 512, 512 and the size of the output layer corresponding to the number
of classes in each dataset. The Adam optimizer is used with a learning rate of
0.001 for all experiments. We train for 5 epochs on MNIST and for 50
epochs on the other datasets.

\minisection{Evaluation protocols.}  In our experiments, we share all
layers in the networks across all tasks during training, which allows
us to perform inference without explicitly knowing the task. We report
the classification accuracy of each previous task $T\!-\!1$ and the
current task $T$ after training on the $T$-th task. When the number of
tasks is large, we only report the average classification accuracy
over all trained tasks.

Lifelong learning is evaluated in two settings depending on knowledge
of the task label at inference time. Here we consider the more
difficult scenario where task labels are unknown, and results cannot
be directly compared to methods which consider
labels~\cite{rebuffi2016icarl,lopez2017gradient,shin2017continual}. As
a consequence our method is implemented with as the last layer in a
single network head, which is also used
in~\cite{chaudhry2018riemannian}. During training we increase the
number of output neurons as new tasks are added.

\begin{table}[tb]
\begin{center}
\caption{Comparison EWC~/~R-EWC for $T\!=\!2$. }
\label{tab:exp1_cifar_vgg}

\begin{tabular}{rcc}
\hline
      &  EWC~\cite{kirkpatrick2017overcoming} (T1 / T2) & R-EWC (T1 / T2) \\ \hline
MNIST & 89.3 (85.8 / 92.8)  & \textbf{93.1} (91.6 / 94.5) \\
CIFAR-100 &37.5 (23.5 / 51.5) &  \textbf{42.5} (30.2 / 54.7)  \\ 
CUB-200 Birds & 45.3 (42.3 / 48.6) &    \textbf{48.4} (53.3 / 45.2) \\ 
Stanford-40 Actions & 50.4 (44.3 / 58.4)  & \textbf{52.5} (52.3 / 52.6) \\ 
\hline
\end{tabular}
\end{center}
\end{table}

\subsection{Disjoint MNIST comparison and ablation study}
We use the disjoint MNIST dataset~\cite{srivastava2013compete}, which
assigns half of the numbers as task 1 and the rest as task 2. We
compare our method (R-EWC) with fine-tuning (FT) and Elastic Weight
Consolidation (EWC)~\cite{kirkpatrick2017overcoming}. First, task 1
is learned from scratch on the LeNet architecture. For FT, task 2 is
learned starting from task 1 as initialization. For EWC and R-EWC,
task 2 is learned according to the corresponding method. In addition,
we also train task 2 with only applying R-EWC to the convolutional
layers (conv only), to fully connected layers (fc only), and to all
layers except for the last fully-connected (all no last).

Table~\ref{tab:exp1_mnist_lenet} compares the performance of the
proposed methods for different values for the trade-off parameter
$\lambda$ between classification loss and FIM regularization. Results
were obtained using 200 randomly selected samples (40 per class) from
the validation set for computing the FIM. Each experiment was executed
3 times and the results in Table~\ref{tab:exp1_mnist_lenet} are the
average. 

Results show that R-EWC clearly outperforms FT and EWC for all
$\lambda$, while the best trade-off value might vary depending on the
layers involved. As expected, lower values of the trade-off tend
towards a more FT strategy where task 1 is forgotten more quickly. On
the other hand, larger values of the trade-off give more importance to
keeping the weights useful for task 1, avoiding catastrophic
forgetting but also making task 2 a bit more difficult to learn. In
conclusion, the improvement of our proposed method over EWC is that
while maintaining similar task 2 performance, it allows for much less
catastrophic forgetting on task 1. We have observed that during
training the regularized part of the FIM is usually between $10^{-2}$
to $10^{-4}$, which could explain why values around $\lambda\!=\!100$
give a more balanced trade-off.

\subsection{Comparison with EWC on two tasks}
We further compare EWC and R-EWC on several larger datasets divided
into two tasks -- that is, in which all datasets are divided into 2
groups with an equal number of classes. The network is trained on task
1 as usual, and then both methods are applied for task 2. After the
learning process is done, we evaluate the two tasks again. The
accuracy for each task and average accuracy of two tasks are reported
in Table~\ref{tab:exp1_cifar_vgg}. Results show that our method
clearly outperforms EWC for all datasets with an absolute gain on
accuracy of R-EWC over EWC that varies from 2.1\% to 5\%. When
comparing the accuracy on the first task only, R-EWC forgets
significantly less in all cases while attaining similar accuracy on
the second task.

\subsection{Comparison with EWC on more tasks}
\begin{figure}[tb]
	\begin{centering}
  \begin{subfigure}[b]{0.75\columnwidth}
    \includegraphics[width=\textwidth]{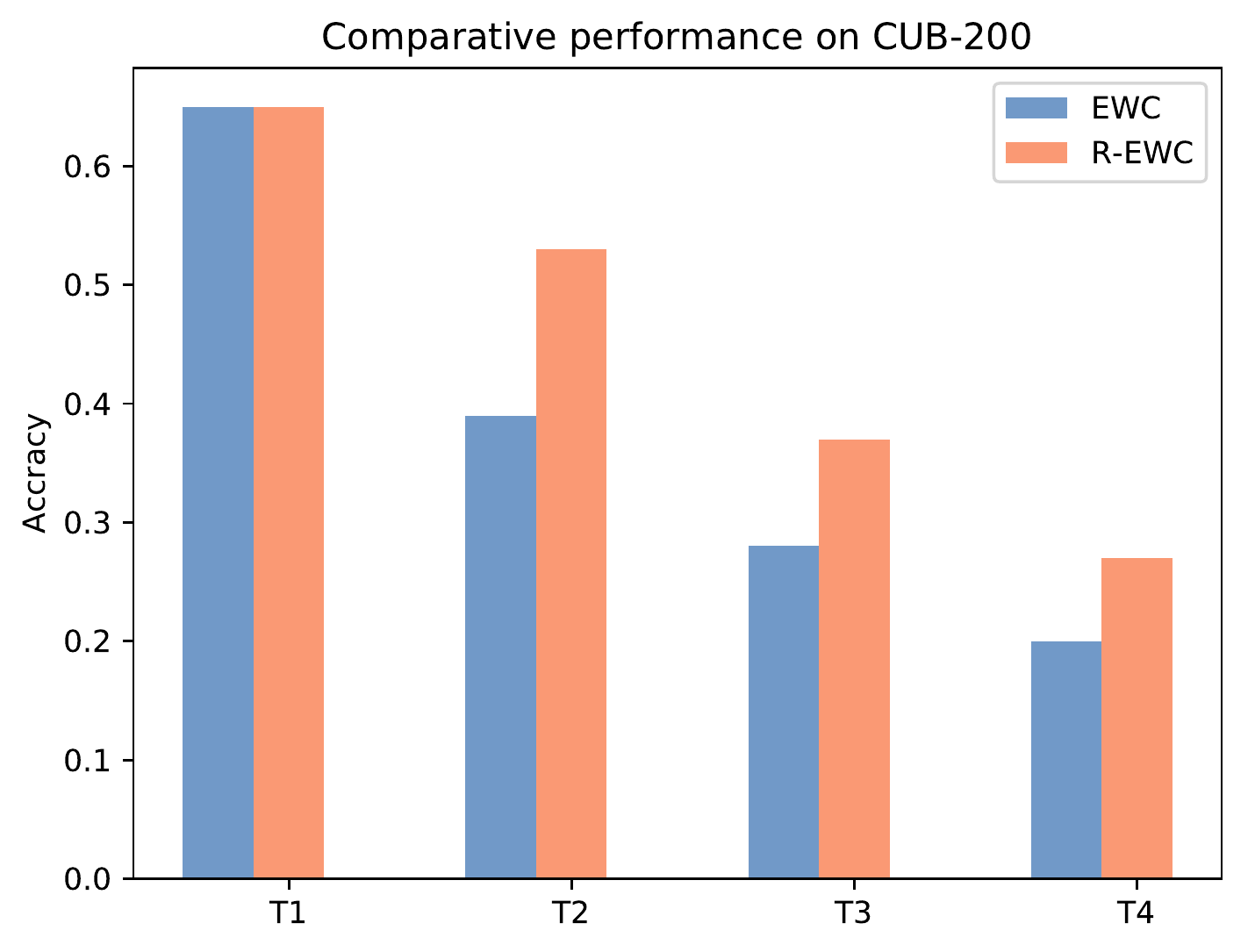}
    \caption{CUB-200 Birds}
    \label{fig:cub}
  \end{subfigure}
    \begin{subfigure}[b]{0.75\columnwidth}
    \includegraphics[width=\textwidth]{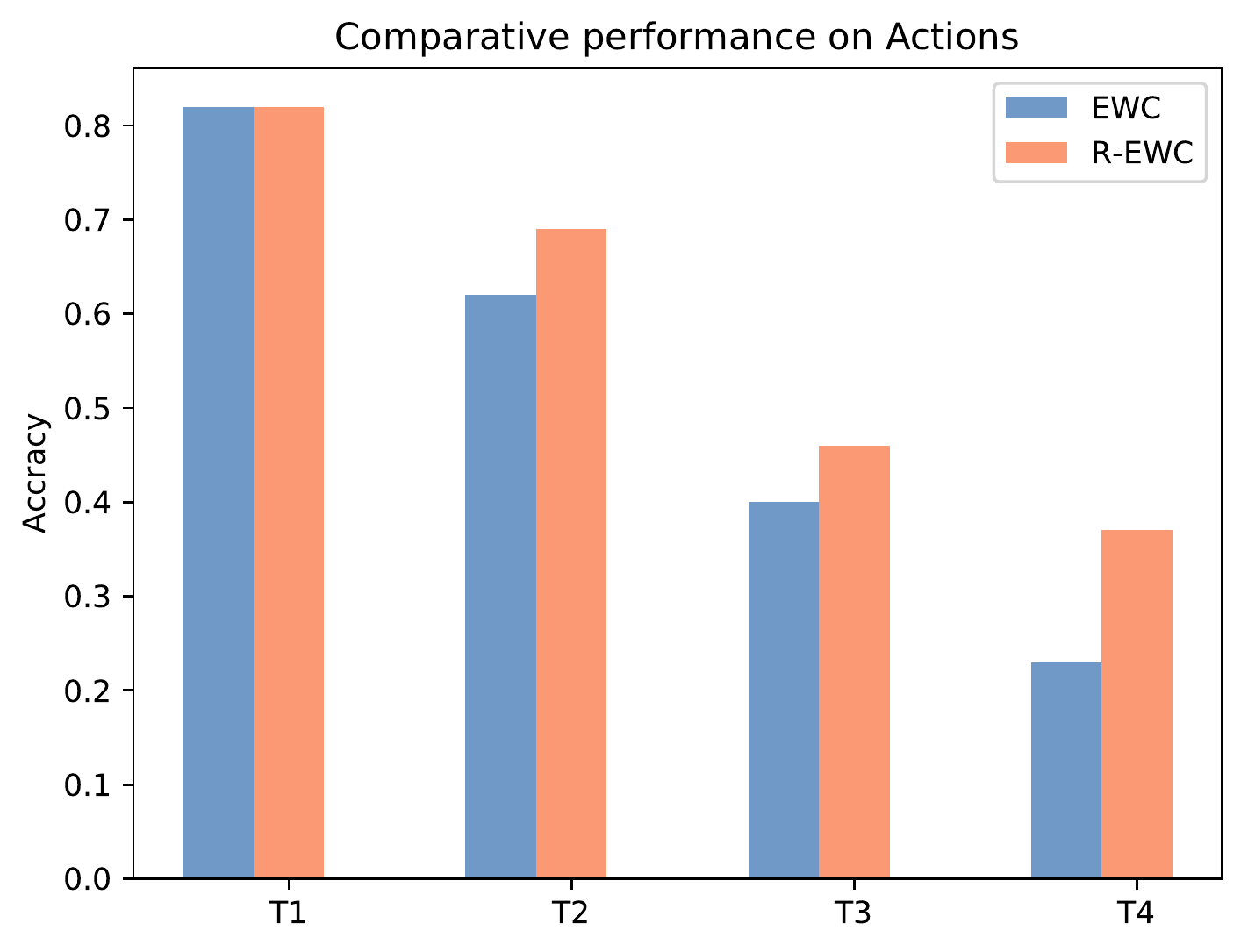}
    \caption{Stanford-40 Actions}
    \label{fig:act}
  \end{subfigure}

    \end{centering}
    \caption{\label{fig:t4} Comparison with EWC when $T\!=\!4$ on CUB-200 Birds and Stanford-40 Actions datasets.}
\end{figure}

We compare both EWC and R-EWC when having more tasks for datasets with
larger images. We divide both CUB-200 Birds and Stanford-40 Actions
datasets into four groups of an equal number of classes. We train a
network on task 1 for both methods and proceed to iteratively learn
the other tasks one at a time, while testing the performance at each
stage. Results are shown in Figure~\ref{fig:t4}, where we observe that
the accuracy decreases with increasing number of tasks for both
methods as expected.\footnote{Fig.~\ref{fig:cub} has been corrected from the original version as it was duplicated.} However, R-EWC outperforms EWC consistently, by a
margin that grows larger as more tasks are learned on Stanford-40 Actions dataset.  Note that in
lifelong learning settings it becomes more difficult to balance
performance of new tasks as the number of previous learned tasks
increases. Results for all previous tasks after training the $T$-th
task on Stanford-40 Actions are given in
Table~\ref{tab:exp1_more_tasks}. For each single previous task, our
method manages to avoid forgetting better than EWC.

\begin{table}[tb]
\centering
\caption{Comparison EWC~/~R-EWC for $T\!=\!4$ on Stanford Actions. }
\label{tab:exp1_more_tasks}
\setlength\tabcolsep{3pt}
\begin{tabular}{cccccc}
\hline
Current & \multicolumn{5}{c}{\begin{tabular}[c]{@{}l@{}}Accuracy \\ \end{tabular}}                                                  \\
Task & T1 & T2 & T3 & T4 & Average\\
\hline
T1 & 81.5 / 81.5 & - & - & -       & 81.5 / 81.5  \\
T2 & 49.5 / \textbf{55.4} & 75.8 / \textbf{81.5} & - & - & 62.0 / \textbf{69.0} \\
T3 & 6.1 / \textbf{18.8} & 45.1 / \textbf{48.7} & 69.9 / \textbf{72.1} & - & 40.3 / \textbf{47.2} \\
T4 & 0.0 / \textbf{12.0} & 7.0 / \textbf{31.3}  & 44.5 / \textbf{56.9} & 46.8 / \textbf{50.9} & 23.0 / \textbf{37.2} \\
\hline
\end{tabular}
\end{table}

\subsection{Comparison with the state-of-the-art}

We compare our method (R-EWC) with fine-tuning (FT), Elastic Weight
Consolidation (EWC)~\cite{kirkpatrick2017overcoming}, Learning without
Forgetting (LwF)~\cite{li2016learning} and Expert Gate
(EG)~\cite{aljundi2016expert}. We exclude methods which use samples of
previous tasks during training of new tasks. We split the CIFAR-100
dataset~\cite{krizhevsky2009learning} into 4 groups of classes, where
each group corresponds to a task. For EG, the base network is trained
on the 4 tasks independently, and for each of them we learn an
auto-encoder with dimensions 4096, 1024 and 100, respectively. For FT,
each task is initialized with the weights of the previous one. In
addition, an UpperBound is shown by learning the newer tasks with all
the images for all previous tasks available.

\begin{figure}
  \centering
  \includegraphics[width=\columnwidth]{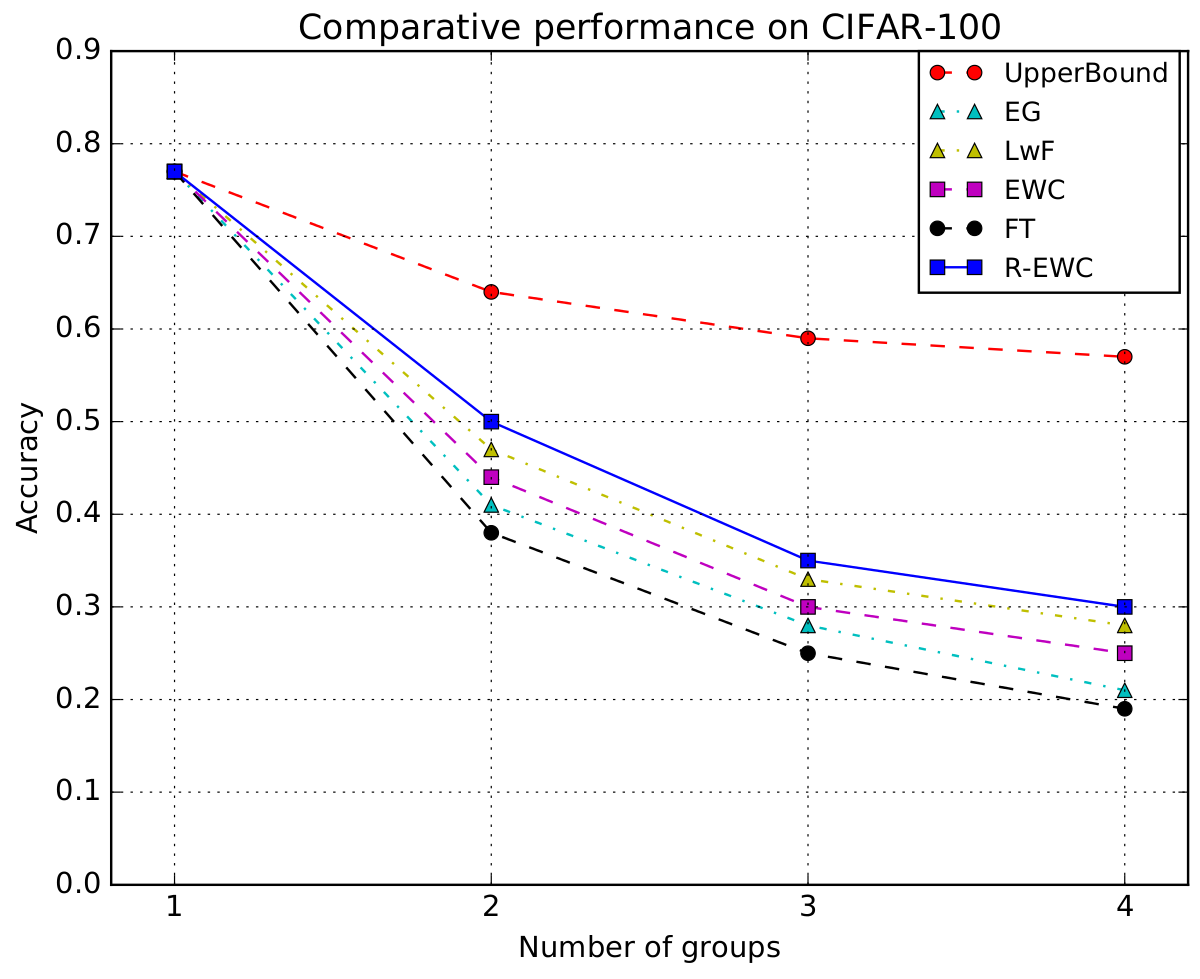}
  \caption{Comparison with the state-of-the-art on CIFAR-100.}
  \label{fig:exp2_cifar100}
\end{figure}

Results are shown in Figure~\ref{fig:exp2_cifar100}, where we clearly
see our method outperforms all others. FT usually performs worse
compared to other baselines, since it tends to forget the previous
tasks completely and is optimal only for the last task. EG usually has
higher accuracy when the tasks are easy to distinguish (on CUB-200 and
Stanford Actions, for example), however it is better than FT but worse
than other baselines in this setting since the groups are randomly
sampled from the CIFAR-100 dataset. EWC performs worse than LwF,
however our method gains about 5\% over EWC and achieves better
performance than LwF. While our method still performs worse than the
UpperBound, this baseline requires all data at all training times and
can not be updated for new tasks.

\section{Conclusions}
\label{sec:conclusions}
EWC helps to prevent forgetting but is very sensitive to the diagonal
approximation of the FIM used in practice (due to the large size of
the full FIM). We show that this approximation discards important
information for preventing forgetting and propose a reparametrization
of the layers that results in more compact and more diagonal FIM. This
reparametrization is based on rotating the FIM in the parameter space
to align it with directions that are less prone to forgetting. Since
direct rotation is not possible due to the feedforward structure of
the network, we devise an indirect method that approximates this
rotation by rotating intermediate features, and that can be easily
implemented as additional convolutional and fully connected
layers. However, the weights in these layers are fixed, so they do not
increase the number of parameters. Our experiments with several tasks
and settings show that EWC in this rotated space (R-EWC) consistently
improves the performance compared to EWC in the original space,
obtaining results that are comparable or better than other
state-of-the-art algorithms using weight consolidation without
exemplars.

\bibliographystyle{IEEEtran}

\minisection{Acknowledgement}
Xialei Liu acknowledges the Chinese Scholarship Council (CSC) grant
No.201506290018. Marc Masana acknowledges 2018-FI\_B1-00198 grant of
\emph{Generalitat de Catalunya}. Luis Herranz acknowledges the
European Union research and innovation program under the Marie
Skłodowska-Curie grant agreement No. 6655919. This work was supported
by TIN2016-79717-R, TIN2017-88709-R, and the CHISTERA project M2CR
(PCIN-2015-251) of the Spanish Ministry, the ACCIO agency and CERCA
Programmes of the \emph{Generalitat de Catalunya}, and the EU Project
CybSpeed MSCA-RISE-2017-777720. We also acknowledge the generous GPU
support from NVIDIA.

\bibliography{refs}
\end{document}